\documentclass[10pt,twocolumn,letterpaper]{article}

\usepackage{iccv}
\usepackage{times}
\usepackage{epsfig}
\usepackage{graphicx}
\usepackage{amsmath}
\usepackage{amssymb}

\graphicspath{{figures/}}

\usepackage[breaklinks=true,colorlinks,bookmarks=false]{hyperref}

\iccvfinalcopy % *** Uncomment this line for the final submission

\def\iccvPaperID{****} % *** Enter the ICCV Paper ID here

\ificcvfinal\fi

\begin{document}

%%%%%%%%% TITLE
\title{Learning Semantic-Specific Graph Representation \\ for Multi-Label Image Recognition}
 \author{Tianshui Chen$^{1,2}$ \and Muxin Xu$^{1}$ \and Xiaolu Hui$^{1}$ \and Hefeng Wu$^{1}$ \thanks{Corresponding author is Hefeng Wu. This work was supported in part by the National Key Research and Development Program of China under Grant No. 2018YFC0830103, in part by National Natural Science Foundation of China (NSFC) under Grant No. 61622214, 61876045, and U1811463, in part by National High Level Talents Special Support Plan (Ten Thousand Talents Program), in part by the Natural Science Foundation of Guangdong Province under Grant No. 2017A030312006, and in part by Zhujiang Science and Technology New Star Project of Guangzhou under Grant No. 201906010057.} \and Liang Lin$^{1,2}$\vspace{1ex}\\
 \and $^1$Sun Yat-sen University ~~~~~~~~~~ $^2$DarkMatter AI Research \\
 {\tt\small \{tianshuichen, wuhefeng\}@gmail.com, \{xumx7, huixlu\}@mail2.sysu.edu.cn, linliang@ieee.org}
}

\maketitle
\ificcvfinal\thispagestyle{empty}\fi

%%%%%%%%% ABSTRACT
\begin{abstract}
Recognizing multiple labels of images is a practical and challenging task, and significant progress has been made by searching semantic-aware regions and modeling label dependency. However, current methods cannot locate the semantic regions accurately due to the lack of part-level supervision or semantic guidance. Moreover, they cannot fully explore the mutual interactions among the semantic regions and do not explicitly model the label co-occurrence. To address these issues, we propose a Semantic-Specific Graph Representation Learning (SSGRL) framework that consists of two crucial modules: 1) a semantic decoupling module that incorporates category semantics to guide learning semantic-specific representations and 2) a semantic interaction module that correlates these representations with a graph built on the statistical label co-occurrence and explores their interactions via a graph propagation mechanism. Extensive experiments on public benchmarks show that our SSGRL framework outperforms current state-of-the-art methods by a sizable margin, e.g. with an mAP improvement of 2.5\%, 2.6\%, 6.7\%, and 3.1\% on the PASCAL VOC 2007 \& 2012, Microsoft-COCO and Visual Genome benchmarks, respectively. Our codes and models are available at \url{https://github.com/HCPLab-SYSU/SSGRL}.
\end{abstract}

%%%%%%%%% BODY TEXT
\section{Introduction}
Multi-label image classification is a fundamental yet practical task in computer vision, as real-world images generally contain multiple diverse semantic objects. Recently, it is receiving increasing attention \cite{li2017improving,wei2016hcp,zhu2017learning}, since it underpins plenty of critical applications in content-based image retrieval and recommendation systems \cite{chua1994concept,yang2015pinterest}. Besides handling the challenges of complex variations in viewpoint, scale, illumination and occlusion, predicting the presence of multiple labels further requires mining semantic object regions as well as modeling the associations and interactions among these regions, rendering multi-label image classification an unsolved and challenging task.

\begin{figure}[!t]
   \centering
   \includegraphics[width=0.95\linewidth]{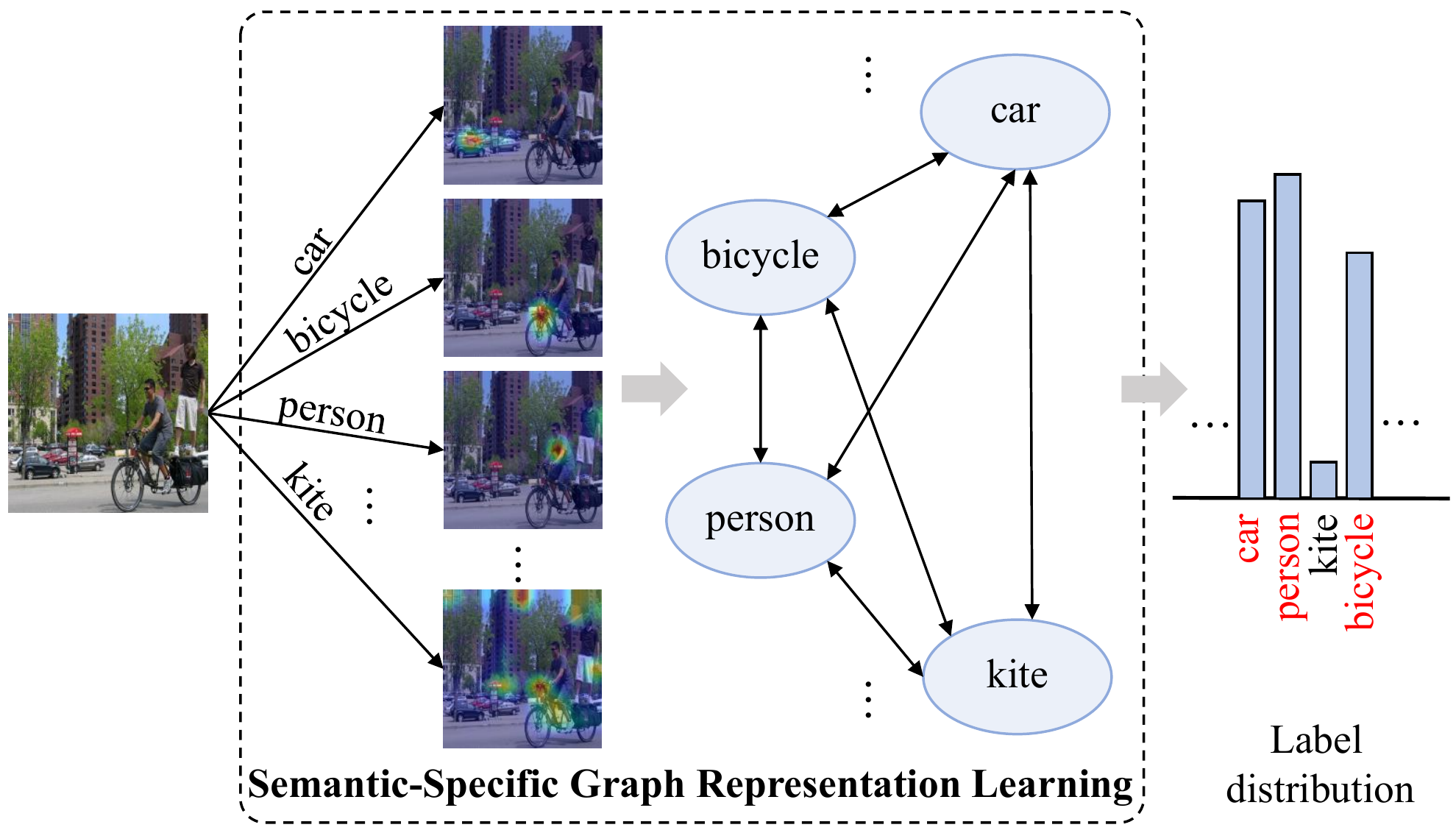} % requires the graphicx package
   \caption{Illustration of our Semantic-Specific Graph Representation Learning framework. It incorporates category semantics to guide learning semantic-specific representation via a semantic decoupling module and explores their interactions via a semantic interaction module.}
   \label{fig:mini-pipeline}
\end{figure}

Current methods for multi-label image classification usually employ object localization techniques \cite{wei2016hcp,yang2016exploit} or resort to visual attention networks \cite{zhu2017learning} to locate semantic object regions. However, object localization techniques \cite{uijlings2013selective,zitnick2014edge} have to search numerous category-agnostic and redundant proposals and can hardly be integrated into deep neural networks for end-to-end training, while visual attention networks can merely locate object regions roughly due to the lack of supervision or guidance. Some other works introduce RNN/LSTM \cite{hochreiter1997long,wang2016cnn,chen2018recurrent} to further model contextual dependencies among semantic regions and capture label dependencies. However, the RNN/LSTM sequentially models regions/labels dependencies, which cannot fully exploit this property since direct association exists between each region or label pair. Besides, they do not explicitly model the statistical label co-occurrence, which is also key to aid multi-label image classification.

To address these issues, we propose a novel Semantic-Specific Graph Representation Learning (SSGRL) framework that incorporates category semantics to guide learning semantic-specific features and explore their interactions to facilitate multi-label image classification. More specifically, we first design a semantic decoupling module that utilizes the semantic features of the categories to guide learning category-related image features that focus more on the corresponding semantic regions (see Figure \ref{fig:mini-pipeline}). Then, we construct a graph based on the statistical label co-occurrence to correlate these features and explore their interactions via a graph propagation mechanism. Figure \ref{fig:mini-pipeline} illustrates a basic pipeline of the proposed SSGRL framework.

The contributions can be summarized into three folds: 1) We formulate a novel Semantic-Specific Graph Representation Learning framework that better learns semantic-specific features and explores their interactions to aid multi-label image recognition. 2) We introduce a novel semantic decoupling module that incorporates category semantics to guide learning semantic-specific features. 3) We conduct experiments on various benchmarks including PASCAL VOC 2007 \& 2012 \cite{everingham2010pascal}, Microsoft-COCO \cite{lin2014microsoft}, and Visual Genome with larger scale categories \cite{krishna2017visual} and demonstrate that our framework exhibits obvious performance improvement. Specifically, it improves the mAP from 92.5\% to 95.0\% and 92.2\% to 94.8\% on the Pascal VOC 2007 and 2012 dataset respectively, from 77.1\% to 83.8\% on the Microsoft-COCO dataset, from 33.5\% to 36.6\% on the Visual Genome 500 dataset compared with current state-of-the-art methods. By simply pre-training on the Microsoft-COCO dataset and fusing two scale results, our framework can further boost the mAP to 95.4\% on the Pascal VOC 2012 dataset.

\section{Related Works}
Recent progress on multi-label image classification relies on the combination of object localization and deep learning techniques \cite{wei2016hcp,yang2016exploit}. Generally, they introduced object proposals \cite{zitnick2014edge} that were assumed to contain all possible foreground objects in the image and aggregated features extracted from all these proposals to incorporate local information. Although these methods achieved notable performance improvement, the step of region candidate localization usually incurred redundant computation cost and prevented the model from end-to-end training with deep neural networks. Zhang et al. \cite{zhang2018multi} further utilized a learning based region proposal network and integrated it with deep neural networks. Although this method could be jointly optimized, it required additional annotations of bounding boxes to train the proposal generation component. To solve this issue, some other works \cite{zhu2017learning,wang2017multi,zhu2017learning} resorted to attention mechanism to locate the informative regions, and these methods could be trained with image level annotations in an end-to-end manner. For example, Wang et al. \cite{wang2017multi} introduced spatial transformer to adaptively search semantic-aware regions and then aggregated features from these regions to identify multiple labels. However, due to the lack of supervision and guidance, these methods could merely locate the regions roughly.

Modeling label dependencies can help capture label co-occurrence, which is also key to aid multi-label recognition. To achieve this, a series of works introduced graphic models, such as Conditional Random Field \cite{ghamrawi2005collective}, Dependency Network \cite{guo2011multi}, or co-occurrence matrix \cite{xue2011correlative} to capture pairwise label correlations. Recently, Wang et al. \cite{wang2016cnn} formulated a CNN-RNN framework that utilized the semantic redundancy and the co-occurrence dependency implicitly to facilitate effective multi-label classification. Some works \cite{zhang2018multi,chen2018recurrent} further took advantage of proposal generation/visual attention mechanism to search local discriminative regions and LSTM \cite{hochreiter1997long} to explicitly model label dependencies. For example, Chen et al. \cite{chen2018recurrent} proposed a recurrent attention reinforcement learning framework to iteratively discover a sequence of attentional and informative regions, and modeled long-term dependencies among these attentional regions that help to capture semantic label co-occurrence. However, the RNN/LSTM \cite{hochreiter1997long} modeled the label dependencies in a sequential manner, and they could not fully exploit the property since mutual dependency might exist between each label pair.

Different from all these methods, our framework incorporates category semantics to guide learning semantic-aware feature vectors. In addition, we directly correlate all label pairs in the form of a structured graph and introduce a graph propagation mechanism to explore their mutual interactions under the explicit guidance of statistical label co-occurrence. Thus, our framework can better learn category-related features and explore their interactions, leading to evident performance improvement.

\section{SSGRL Framework}
\subsection{Overview}
\begin{figure*}[!t]
   \centering
   \includegraphics[width=0.95\linewidth]{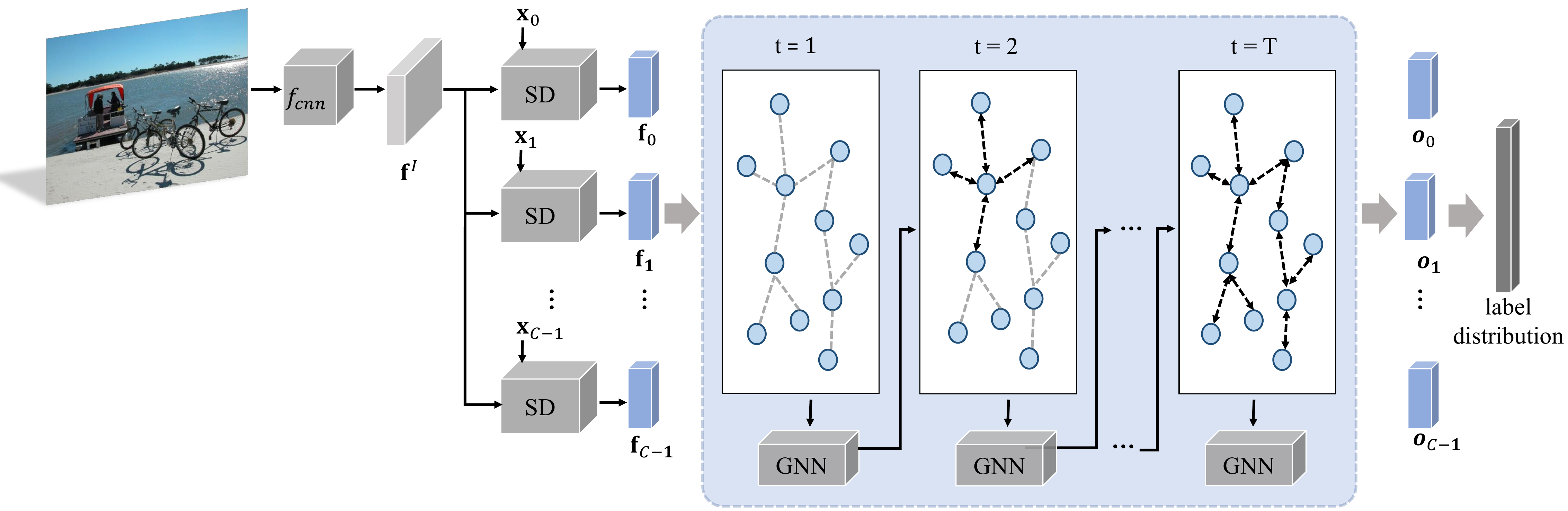} % requires the graphicx package
   \caption{Illustration of our Semantic-Specific Graph Representation Learning framework. Given an input image, we first feed it into a CNN to extract image representation. Then, a semantic decoupling (SD) module incorporates category semantics to guide learning semantic-specific representations, and a semantic interaction module correlates these representations using a graph and adopts a graph neural network (GNN) to explore their interactions.}
   \label{fig:pipeline}
\end{figure*}

In this section, we first give an overall description of the proposed SSGRL framework that consists of two crucial modules, i.e, semantic decoupling and semantic interaction. Given an image, we first feed it into a fully convolutional network to generate its feature maps. Then, for each category, the semantic decoupling module incorporates the category semantics to guide learning semantic-specific representations that focus on the semantic regions of this category. Finally, the semantic interaction module correlates these representations using a graph that is constructed based on the statistical label co-occurrence, and it explores the semantic interactions using a graph propagation network to further learn contextualized features, which are then used to predict the final label distribution. Figure \ref{fig:pipeline} illustrates a detailed pipeline of the SSGRL framework.

\subsection{Semantic Decoupling}
The semantic decoupling module aims to learn semantic-specific feature representation by taking the category semantics as guidance. Here, we adopt a semantic guided attention mechanism to implement this module.

Given an input image $I$, the framework first extracts its feature maps $\mathbf{f}^I \in \mathcal{R}^{W\times H \times N}$, where $W$, $H$, and $N$ are the width, height and channel number of the feature maps, formulated as
\begin{equation}
  \mathbf{f}^I=f_{cnn}(I),
  \label{eq:image-feature-extraction}
\end{equation}
where $f_{cnn}(\cdot)$ is a feature extractor, and it is implemented by a fully convolutional network. For each category $c$, the framework extracts a $d_s$-dimensional semantic-embedding vector using the pre-trained GloVe \cite{pennington2014glove} model
\begin{equation}
  \mathbf{x}_c=f_{g}(w_c),
  \label{eq:semantic-feature-extraction}
\end{equation}
where $w_c$ is the semantic word of category $c$. Then, we introduce a semantic guided attention mechanism which incorporates the semantic vector $\mathbf{x}_c$ to guide focusing more on the semantic-aware regions and thus learning a feature vector corresponding to this category. More specifically, for each location $(w, h)$, we first fuse the corresponding image feature $\mathbf{f}^I_{wh}$ and $\mathbf{x}_c$ using a low-rank bilinear pooling method \cite{kim2016hadamard}
\begin{equation}
  \tilde{\mathbf{f}}^I_{c,wh}=\mathbf{P}^T\left(\tanh\left((\mathbf{U}^T\mathbf{f}^I_{wh})\odot(\mathbf{V}^T\mathbf{x}_c)\right)\right)+\mathbf{b},
\end{equation}
where $\tanh(\cdot)$ is the hyperbolic tangent function, $\mathbf{U}\in \mathcal{R}^{N\times d_1}$, $\mathbf{V}\in \mathcal{R}^{d_s\times d_1}$, $\mathbf{P}\in \mathcal{R}^{d_1\times d_2}$, $\mathbf{b}\in \mathcal{R}^{d_2}$ are the learnable parameters, and $\odot$ is the element-wise multiplication operation. $d_1$ and $d_2$ are the dimensions of the joint embeddings and the output features. Then, an attentional coefficient is computed under the guidance of $\mathbf{x}_c$ by
\begin{equation}
  \tilde{a}_{c,wh}=f_{a}(\tilde{\mathbf{f}}^I_{c,wh}).
\end{equation}
This coefficient indicates the importance of location $(w, h)$. $f_{a}(\cdot)$ is an attentional function and it is implemented by a fully connected network. The process is repeated for all locations. To make the coefficients easily comparable across different samples, we normalize the coefficients over all locations using a softmax function
\begin{equation}
  a_{c,wh}=\frac{\mathrm{exp}(\tilde{a}_{c,wh})}{\sum_{w',h'}{\mathrm{exp}(\tilde{a}_{c,w'h'}})}.
\end{equation}
Finally, we perform weighted average pooling over all locations to obtain a feature vector
\begin{equation}
  \mathbf{f}_{c}=\sum_{w,h}a_{c,wh}\mathbf{f}_{c,wh}
\end{equation}
that encodes the information related to category $c$. We repeat the process for all categories and obtain all the category-related feature vectors $\{\mathbf{f}_{0}, \mathbf{f}_{1}, \dots, \mathbf{f}_{C-1}\}$.

\subsection{Semantic Interaction}
\label{sec:SI}
Once obtaining the feature vectors corresponding to all categories, we correlate these vectors in the form of a graph that is constructed based on the statistical label co-occurrence and introduce a graph neural network to propagate message through the graph to explore their interactions.

\noindent\textbf{Graph construction.} We first introduce the graph $\mathcal{G}=\{\mathbf{V}, \mathbf{A}\}$, in which nodes refer to the categories and edges refer to the co-occurrence between corresponding categories. Specifically, suppose that the dataset covers $C$ categories, $\mathbf{V}$ can be represented as $\{v_0, v_2, \dots, v_{C-1}\}$ with element $v_c$ denoting category $c$ and $\mathbf{A}$ can be represented as $\{a_{00}, a_{01}, \dots, a_{0(C-1)}, \dots, a_{(C-1)(C-1)}\}$ with element $a_{cc'}$ denoting the probability of the existence of object belonging to category $c'$ in the presence of object belonging to category $c$. We compute the probabilities between all category pairs using the label annotations of samples on the training set, thus we do not introduce any additional annotation.

Inspired by the current graph propagation works \cite{li2015gated,chen2018knowledge,wang2018deep,chen2019knowledge}, we adopt a gated recurrent update mechanism to propagate message through the graph and learn contextualized node-level features. Specifically, for each node $v_c \in \mathbf{V}$, it has a hidden state $\mathbf{h}_c^t$ at timestep $t$. In this work, as each node corresponds to a specific category and our model aims to explore the interactions among the semantic-specific features, we initialize the hidden state at $t=0$ with the feature vector that relates to the corresponding category, formulated as
\begin{equation}
  \mathbf{h}_c^0=\mathbf{f}_{c}.
\end{equation}
At timestep $t$, the framework aggregates message from its neighbor nodes, expressed as
\begin{equation}
  \mathbf{a}_c^t=\left[\sum_{c'}(a_{cc'})\mathbf{h}_c^{t-1}, \sum_{c'}(a_{c'c})\mathbf{h}_c^{t-1}\right].
\end{equation}
In this way, the framework encourages message propagation if node $c'$ has a high correlation with node $c$, and it suppresses propagation otherwise. Therefore, it can propagate message through the graph and explore node interactions under the guidance of the prior knowledge of statistical label co-occurrence. Then, the framework updates the hidden state based on the aggregated feature vector $\mathbf{a}_c^t$ and its hidden state at previous timestep $\mathbf{h}_c^{t-1}$ via a gated mechanism similar to the Gated Recurrent Unit, formulated as
\begin{equation}
   \begin{split}
    \mathbf{z}_c^t=&{}\sigma(\mathbf{W}^z{\mathbf{a}_c^t}+\mathbf{U}^z{\mathbf{h}_c^{t-1}}) \\
    \mathbf{r}_c^t=&{}\sigma(\mathbf{W}^r{\mathbf{a}_c^t}+\mathbf{U}^r{\mathbf{h}_c^{t-1}}) \\
    \widetilde{\mathbf{h}_c^t}=&{}\tanh\left(\mathbf{W}{\mathbf{a}_c^t}+\mathbf{U}({\mathbf{r}_c^t}\odot{\mathbf{h}_c^{t-1}})\right) \\
    \mathbf{h}_c^t=&{}(1-{\mathbf{z}_c^t}) \odot{\mathbf{h}_c^{t-1}}+{\mathbf{z}_c^t}\odot{\widetilde{\mathbf{h}_c^t}}
   \end{split}
   \label{eq:ggnn}
\end{equation}
where $\sigma(\cdot)$ is the logistic sigmoid function, $\tanh(\cdot)$ is the hyperbolic tangent function, and $\odot$ is the element-wise multiplication operation. In this way, each node can aggregate message from other nodes and simultaneously transfer its information through the graph, enabling interactions among all feature vectors corresponding to all categories. The process is repeated $T$ times, and the final hidden states are generated, i.e., $\{\mathbf{h}_0^T, \mathbf{h}_1^T, \dots, \mathbf{h}_{C-1}^T\}$. Here, the hidden state of each node $\mathbf{h}_{c}^T$ not only encodes features of category $c$, but also carries contextualized message from other categories. Finally, we concatenate $\mathbf{h}_{c}^T$ and the input feature vector $\mathbf{h}_{c}^0$ to predict the confidence score of the presence of category $c$, formulated as
\begin{equation}
   \begin{split}
    \mathbf{o}_c=&f_o(\mathbf{h}_{c}^T, \mathbf{h}_{c}^0) \\
    s_c=&f_c(\mathbf{o}_c) \\
   \end{split}
   \label{eq:cls}
\end{equation}
where $f_o(\cdot)$ is an output function that maps the concatenation of $\mathbf{h}_{c}^T$ and $\mathbf{h}_{c}^0$ into an output vector $\mathbf{o}_c$. We adopt $C$ classification functions with unshared parameters $\{f_0, f_1, \dots, f_{C-1}\}$, in which $f_c(\cdot)$ takes $\mathbf{o}_c$ as input to predict a score to indicate the probability of category $c$. We perform the process for all categories and obtain a score vector $\mathbf{s}=\{s_0, s_1, \dots, s_{C-1}\}$.

\subsection{Network Architecture}
Following existing multi-label image classification works \cite{zhu2017learning}, we implement the feature extractor $f_{cnn}(\cdot)$ based on the widely used ResNet-101 \cite{he2016deep}. Specifically, we replace the last average pooling layer with another average pooling layer with a size of $2\times 2$ and a stride of 2, with other layers unchanged for implementation. For the low rank bilinear pooling operation, $N$, $d_s$, $d_1$, and $d_2$ are set as 2,048, 300, 1,024, and 1,024, respectively. Thus, $f_a(\cdot)$ is implemented by a 1,024-to-1 fully connected layer that maps the 1,024 feature vector to one single attentional coefficient.

For the graph neural network, we set the dimension of the hidden state as 2,048 and the iteration number $T$ as 3. The dimension of output vector $\mathbf{o}_c$ is also set as 2,048. Thus, the output network $o(\cdot)$ can be implemented by a 4,096-to-2,048 fully connected layer followed by the hyperbolic tangent function, and each classification network $f_c(\cdot)$ can be implemented by a 2,048-to-1 fully connected layer.

\begin{table*}[!t]
\centering
\begin{tabular}{c|c|cccccc|cccccc}
\hline
&  & \multicolumn{6}{c|}{Top 3} &\multicolumn{6}{c}{All} \\
\hline
\centering  Methods & mAP & CP & CR & CF1 & OP & OR & OF1  & CP & CR & CF1 & OP & OR & OF1 \\
\hline
\hline
WARP \cite{gong2013deep}   &- & 59.3 & 52.5 & 55.7 & 59.8 & 61.4 & 60.7 & - & - & -&-  & -&- \\
CNN-RNN \cite{wang2016cnn} &- & 66.0 & 55.6 & 60.4 & 69.2 & 66.4 & 67.8 &- & - & -&-  & -&-\\
RLSD \cite{zhang2018multi} &- & 67.6 & 57.2 & 62.0 & 70.1 & 63.4 & 66.5 &- & - & -&-  & -&-\\
RARL \cite{chen2018recurrent} & - & 78.8 & 57.2 & 66.2 & 84.0 & 61.6 & 71.1 & \\
RDAR \cite{wang2017multi} & 73.4 & 79.1 &  58.7 & 67.4 & 84.0 & 63.0 & 72.0 &- & - & -&-  & -&-\\
KD-WSD \cite{DBLP:conf/mm/LiuSSYXP18}& 74.6 & - & -& 66.8 & - &- &72.7  &- &-& 69.2 & - & -&74.0 \\
ResNet-SRN-att \cite{zhu2017learning} & 76.1 & 85.8 & 57.5 & 66.3 & 88.1 & 61.1 & 72.1 & 81.2 & 63.3 & 70.0 & 84.1 & 67.7 & 75.0\\
ResNet-SRN \cite{zhu2017learning} & 77.1 & 85.2 & 58.8 & 67.4 & 87.4 & 62.5 & 72.9 & 81.6 & 65.4 & 71.2 & 82.7 & 69.9 & 75.8 \\
\hline
Ours & \textbf{83.8} & \textbf{91.9} & \textbf{62.5} & \textbf{72.7} & \textbf{93.8} & \textbf{64.1} & \textbf{76.2} & \textbf{89.9} & \textbf{68.5} & \textbf{76.8} & \textbf{91.3} & \textbf{70.8} & \textbf{79.7}\\
\hline
\end{tabular}
\caption{Comparison of mAP, CP, CR, CF1 and OP, OR, OF1 (in \%) of our framework and state-of-the-art methods under the settings of all and top-3 labels on the Microsoft COCO dataset. ``-'' denotes the corresponding result is not provided.}
\vspace{-10pt}
\label{table:result-coco}
\end{table*}

\subsection{Optimization}
\label{sec:optimization}
Given a dataset that contains $M$ training samples $\{I_i, y_i\}_{i=0}^{M-1}$, in which $I_i$ is the $i$-th image and $y_i=\{y_{i0}, y_{i1}, \dots, y_{i(C-1)}\}$ is the corresponding annotation. $y_{ic}$ is assigned as 1 if the sample is annotated with category $c$ and 0 otherwise. Given an image $I_i$, we can obtain a predicted score vector $\mathbf{s}_i=\{s_{i0}, s_{i1}, \dots, s_{i(C-1)}\}$ and compute the corresponding probability vector $\mathbf{p}_i=\{p_{i0}, p_{i1}, \dots, p_{i(C-1)}\}$ via a sigmoid function
\begin{equation}
  p_{ic}=\sigma(s_{ic}).
\end{equation}
We adopt the cross entropy as the objective loss function
\begin{equation}
  \mathcal{L}=\sum_{i=0}^{N-1}\sum_{c=0}^{C-1}\left(y_{ic}\log p_{ic}+(1-y_{ic})\log(1-p_{ic})\right).
\end{equation}

The proposed framework is trained with the loss $\mathcal{L}$ in an end-to-end fashion. Specifically, we first utilize the ResNet-101 parameters pre-trained on the ImageNet dataset \cite{deng2009imagenet} to initialize the parameters of the corresponding layers in $f_{cnn}$ and initialize the parameters of other layers randomly. As the lower layers' parameters pre-trained on the ImageNet dataset generalize well across different datasets, we fix the parameters of the previous 92 convolutional layers in $f_{cnn}(\cdot)$ and jointly optimize all the other layers. The framework is trained with ADAM algorithm \cite{kingma2014adam} with a batch size of 4, momentums of 0.999 and 0.9. The learning rate is initialized as $10^{-5}$ and it is divided by 10 when the error plateaus. During training, the input image is resized to $640\times 640$, and we randomly choose a number from $\{640, 576, 512, 384, 320\}$ as the width and height to randomly crop patches. Finally, the cropped patches are further resized to $576\times 576$. During testing, we simply resize the input image to $640\times 640$ and perform center crop with a size of $576\times 576$ for evaluation.

\section{Experiments}

\subsection{Evaluation Metrics}
To fairly compare with existing methods, we follow them to adopt the average precision (AP) on each category and mean average precision (mAP) over all categories for evaluation \cite{wei2016hcp,yang2016exploit}. We also follow previous works \cite{zhu2017learning,li2017improving} to present the precision, recall, and F1-measure for further comparison. Here, we assign the labels with top-$3$ highest scores for each image and compare them with the ground truth labels. Concretely, we adopt the overall precision, recall, F1-measure (OP, OR, OF1) and per-class precision, recall, F1-measure (CP, CR, CF1), which are defined as below
\begin{small}
\begin{equation}
   \begin{split}
         \mathrm{OP}&=\frac{\sum_{i}N_{i}^{c}}{\sum_{i}N_{i}^{p}},\quad\\
         \mathrm{OR}&=\frac{\sum_{i}N_{i}^{c}}{\sum_{i}N_{i}^{g}},\quad\\
         \mathrm{OF}1&=\frac{2 \times \mathrm{OP} \times \mathrm{OR}}{\mathrm{OP}+\mathrm{OR}},\quad
   \end{split}
      \begin{split}
         \mathrm{CP}&=\frac{1}{C}\sum_{i}\frac{N_{i}^{c}}{N_{i}^{p}}\\
         \mathrm{CR}&=\frac{1}{C}\sum_{i}\frac{N_{i}^{c}}{N_{i}^{g}}\\
         \mathrm{CF}1&=\frac{2 \times \mathrm{CP} \times \mathrm{CR}}{\mathrm{CP}+\mathrm{CR}}
   \end{split}
   \label{eqn:metric}
\end{equation}
\end{small}%
where $C$ is the number of labels, $N_{i}^{c}$ is the number of images that are correctly predicted for the $i$-th label, $N_{i}^{p}$ is the number of predicted images for the $i$-th label, $N_{i}^{g}$ is the number of ground truth images for the $i$-th label. The above metrics require a fixed number of labels, but the label numbers of different images are generally various. Thus, we further present the OP, OR, OF1 and CP, CR, CF1 metrics under the setting that a label is predicted as positive if its estimated probability is greater than 0.5 \cite{zhu2017learning}. Among these metrics, mAP, OF1, and CF1 are the most important metrics that can provide a more comprehensive evaluation.

\begin{table*}[htp]
\centering
\scriptsize
\begin{tabular}
{p{2.2cm}|p{0.2cm}p{0.2cm}p{0.2cm}p{0.2cm}p{0.3cm}p{0.3cm}p{0.20cm}p{0.20cm}p{0.3cm}p{0.3cm}p{0.3cm}p{0.3cm}p{0.3cm}p{0.4cm}p{0.3cm}p{0.3cm}p{0.3cm}p{0.3cm}p{0.3cm}p{0.3cm}|p{0.5cm}}
\hline
\centering Methods  & aero & bike & bird & boat & bottle & bus & car & cat & chair & cow & table & dog & horse & mbike & person & plant & sheep & sofa & train & tv & mAP \\
\hline
\hline
\centering CNN-RNN~\cite{wang2016cnn} & 96.7 & 83.1 & 94.2 & 92.8 & 61.2 & 82.1 & 89.1 & 94.2 & 64.2 & 83.6 & 70.0 & 92.4 & 91.7 & 84.2 & 93.7 & 59.8 & 93.2 & 75.3 & 99.7 & 78.6 & 84.0 \\
\centering RMIC \cite{DBLP:conf/aaai/He0G0T18} & 97.1 &  91.3 & 94.2 &  57.1 &  86.7 & 90.7 &  93.1 & 63.3 & 83.3 &  76.4 & 92.8 & 94.4 & 91.6 & 95.1 & 92.3 & 59.7 & 86.0 & 69.5 & 96.4 & 79.0 & 84.5 \\
\centering VGG16+SVM~\cite{simonyan2014very}   & - & -&-  & -&- & - & -&-  &- & -&  -& -& - & -& -&-  &- & - &- &- & 89.3\\
\centering VGG19+SVM~\cite{simonyan2014very}   & - & -&-  & -&- & - & -&-  &- & -&  -& -& - & -& -&-  &- & - &- &-  & 89.3\\
\centering RLSD \cite{zhang2016multi} & 96.4 &  92.7 & 93.8 & 94.1 & 71.2 &  92.5 &  94.2 & 95.7 & 74.3 &  90.0 & 74.2  & 95.4 & 96.2 &  92.1 & 97.9 & 66.9 & 93.5 & 73.7 & 97.5 & 87.6 & 88.5\\
\centering HCP~\cite{wei2016hcp}  & 98.6 & 97.1  & \textcolor[rgb]{1,0,0}{98.0} & 95.6 & 75.3&94.7  & 95.8 &97.3 & 73.1 & 90.2 & 80.0 & 97.3 & 96.1 & 94.9 & 96.3 & 78.3 & 94.7 & 76.2 & 97.9 & 91.5 & 90.9\\
\centering FeV+LV~\cite{yang2016exploit}  & 97.9 & 97.0 & 96.6 & 94.6 & 73.6 & 93.9 & 96.5& 95.5 & 73.7 & 90.3 & 82.8 & 95.4 & 97.7 & 95.9 & 98.6 & 77.6 & 88.7 & 78.0 & 98.3 & 89.0 & 90.6\\
\centering RDAR \cite{wang2017multi}  & 98.6 & 97.4 & 96.3 & 96.2 & 75.2 & 92.4 & 96.5 & 97.1 & 76.5 & 92.0 & \textcolor[rgb]{0,0,1}{87.7} & 96.8 & 97.5  & 93.8 & 98.5 & 81.6 & 93.7 & \textcolor[rgb]{0,0,1}{82.8} & \textcolor[rgb]{0,0,1}{98.6} &89.3 & 91.9 \\
\centering RARL \cite{chen2018recurrent} & 98.6 & 97.1 & 97.1 & 95.5 & 75.6 & 92.8 & \textcolor[rgb]{0,0,1}{96.8} & 97.3 & \textcolor[rgb]{0,0,1}{78.3} & 92.2 & 87.6 & 96.9 & 96.5 & 93.6 & 98.5 & 81.6 & 93.1 & 83.2 & 98.5 & 89.3 & 92.0 \\
\centering RCP \cite{wang2016beyond} & 99.3 & \textcolor[rgb]{0,0,1}{97.6} & \textcolor[rgb]{1,0,0}{98.0} & 96.4 & 79.3 & 93.8 & 96.6 & 97.1 & 78.0 & 88.7 & 87.1 & 97.1 & 96.3 & 95.4 & \textcolor[rgb]{1,0,0}{99.1} & 82.1 & 93.6 & 82.2 & 98.4 & \textcolor[rgb]{0,0,1}{92.8} & 92.5 \\
\centering \textbf{Ours} & \textcolor[rgb]{0,0,1}{99.5} & 97.1  & \textcolor[rgb]{0,0,1}{97.6} &  \textcolor[rgb]{1,0,0}{97.8} & \textcolor[rgb]{0,0,1}{82.6} & \textcolor[rgb]{0,0,1}{94.8} & 96.7 & \textcolor[rgb]{0,0,1}{98.1} & 78.0 &\textcolor[rgb]{0,0,1}{97.0} & 85.6 & \textcolor[rgb]{0,0,1}{97.8} & \textcolor[rgb]{0,0,1}{98.3} & \textcolor[rgb]{0,0,1}{96.4} & 98.8 & \textcolor[rgb]{0,0,1}{84.9} & \textcolor[rgb]{0,0,1}{96.5} & 79.8 & 98.4 & \textcolor[rgb]{0,0,1}{92.8} & \textcolor[rgb]{0,0,1}{93.4}\\
\centering \textbf{Ours (pre)} & \textcolor[rgb]{1,0,0}{99.7} & \textcolor[rgb]{1,0,0}{98.4} & \textcolor[rgb]{1,0,0}{98.0} & \textcolor[rgb]{0,0,1}{97.6} & \textcolor[rgb]{1,0,0}{85.7} & \textcolor[rgb]{1,0,0}{96.2} & \textcolor[rgb]{1,0,0}{98.2} & \textcolor[rgb]{1,0,0}{98.8} & \textcolor[rgb]{1,0,0}{82.0}  & \textcolor[rgb]{1,0,0}{98.1} & \textcolor[rgb]{1,0,0}{89.7} & \textcolor[rgb]{1,0,0}{98.8} & \textcolor[rgb]{1,0,0}{98.7} & \textcolor[rgb]{1,0,0}{97.0} & \textcolor[rgb]{0,0,1}{99.0} & \textcolor[rgb]{1,0,0}{86.9} & \textcolor[rgb]{1,0,0}{98.1} & \textcolor[rgb]{1,0,0}{85.8} & \textcolor[rgb]{1,0,0}{99.0} & \textcolor[rgb]{1,0,0}{93.7} & \textcolor[rgb]{1,0,0}{95.0} \\
\hline
\hline
\centering VGG16\&19+SVM~\cite{simonyan2014very}  & 98.9 & 95.0 & 96.8 & 95.4 & 69.7 & 90.4 & 93.5 & 96.0 & 74.2 & 86.6 & 87.8 & 96.0 & 96.3 & 93.1 & 97.2 & 70.0 & 92.1 & 80.3 & 98.1 & 87.0 & 89.7\\
\centering FeV+LV (fusion)~\cite{yang2016exploit} & 98.2 & 96.9 & 97.1 & 95.8 & 74.3 & 94.2 & 96.7 & 96.7 & 76.7 & 90.5 & 88.0 & 96.9 & 97.7 & 95.9 & 98.6 & 78.5 & 93.6 & 82.4 & 98.4 & 90.4 & 92.0 \\
\hline
\end{tabular}
\vspace{1pt}
\caption{Comparison of AP and mAP in \% of our framework and state-of-the-art methods on the PASCAL VOC 2007 dataset. Upper part presents the results of single model and lower part presents those that aggregate multiple models.
``Ours'' and ``Ours (pre)'' denote our framework without and with pre-training on the COCO dataset.
The best and second best results are highlighted in {\color{red}{red}} and {\color{blue}{blue}}, respectively. ``-'' denotes the corresponding result is not provided. Best viewed in color.}
\label{table:comparision_voc07}
\end{table*}

\begin{table*}[htp]
\centering
\scriptsize
\begin{tabular}
{p{2.2cm}|p{0.2cm}p{0.2cm}p{0.2cm}p{0.2cm}p{0.3cm}p{0.3cm}p{0.20cm}p{0.20cm}p{0.3cm}p{0.3cm}p{0.3cm}p{0.3cm}p{0.3cm}p{0.4cm}p{0.3cm}p{0.3cm}p{0.3cm}p{0.3cm}p{0.3cm}p{0.3cm}|p{0.5cm}}
\hline
\centering Methods  & aero & bike & bird & boat & bottle & bus & car & cat & chair & cow & table & dog & horse & mbike & person & plant & sheep & sofa & train & tv & mAP \\
\hline
\hline
\centering RMIC  \cite{DBLP:conf/aaai/He0G0T18} & 98.0 & 85.5 & 92.6 & 88.7 & 64.0 & 86.8 & 82.0 & 94.9 & 72.7 & 83.1 & 73.4 & 95.2 & 91.7 & 90.8 &95.5 & 58.3 & 87.6 & 70.6 & 93.8 & 83.0 & 84.4 \\
\centering VGG16+SVM~\cite{simonyan2014very} & 99.0 & 88.8 & 95.9 & 93.8 & 73.1 & 92.1 & 85.1 & 97.8 & 79.5 & 91.1 & 83.3 & 97.2 & 96.3 & 94.5 & 96.9 & 63.1 & 93.4 & 75.0 & 97.1 & 87.1 & 89.0 \\
\centering VGG19+SVM~\cite{simonyan2014very} & 99.1 & 88.7 & 95.7 & 93.9 & 73.1 & 92.1 & 84.8 & 97.7 & 79.1 & 90.7 & 83.2 & 97.3 & 96.2 & 94.3 & 96.9 & 63.4 & 93.2 & 74.6 & 97.3 & 87.9 & 89.0 \\
\centering HCP \cite{wei2016hcp} & 99.1 & 92.8 & 97.4 & 94.4 & 79.9 & 93.6 & 89.8 & 98.2 & 78.2 & 94.9 & 79.8 & 97.8 & 97.0 & 93.8 & 96.4 & 74.3 & 94.7 & 71.9 & 96.7 & 88.6 & 90.5 \\
\centering FeV+LV~\cite{yang2016exploit} & 98.4 & 92.8 & 93.4 & 90.7 & 74.9 & 93.2 & 90.2 & 96.1 & 78.2 & 89.8 & 80.6 & 95.7 & 96.1 & 95.3 & 97.5 & 73.1 & 91.2 & 75.4 & 97.0 & 88.2 & 89.4 \\
\centering RCP \cite{wang2016beyond} & 99.3 & 92.2 & \textcolor[rgb]{0,0,1}{97.5} & 94.9 & 82.3 & 94.1 & 92.4 & \textcolor[rgb]{0,0,1}{98.5} & 83.8 & 93.5 & 83.1 & 98.1 & 97.3 & 96.0 & \textcolor[rgb]{0,0,1}{98.8} & 77.7 & 95.1 & 79.4 & 97.7 & 92.4 & 92.2 \\
\centering \textbf{Ours}  & \textcolor[rgb]{0,0,1}{99.5} & \textcolor[rgb]{0,0,1}{95.1} & 97.4 & \textcolor[rgb]{0,0,1}{96.4} & \textcolor[rgb]{0,0,1}{85.8} & \textcolor[rgb]{0,0,1}{94.5} & \textcolor[rgb]{0,0,1}{93.7} & \textcolor[rgb]{1,0,0}{98.9} & \textcolor[rgb]{0,0,1}{86.7} & \textcolor[rgb]{0,0,1}{96.3} & \textcolor[rgb]{0,0,1}{84.6} & \textcolor[rgb]{0,0,1}{98.9} & \textcolor[rgb]{0,0,1}{98.6} &  \textcolor[rgb]{0,0,1}{96.2} & 98.7 & \textcolor[rgb]{0,0,1}{82.2}  & \textcolor[rgb]{0,0,1}{98.2} & \textcolor[rgb]{0,0,1}{84.2} & \textcolor[rgb]{0,0,1}{98.1} & \textcolor[rgb]{0,0,1}{93.5} & \textcolor[rgb]{0,0,1}{93.9} \\
\centering \textbf{Ours (pre)} & \textcolor[rgb]{1,0,0}{99.7} & \textcolor[rgb]{1,0,0}{96.1} & \textcolor[rgb]{1,0,0}{97.7} & \textcolor[rgb]{1,0,0}{96.5} & \textcolor[rgb]{1,0,0}{86.9} & \textcolor[rgb]{1,0,0}{95.8} & \textcolor[rgb]{1,0,0}{95.0} & \textcolor[rgb]{1,0,0}{98.9} & \textcolor[rgb]{1,0,0}{88.3} & \textcolor[rgb]{1,0,0}{97.6} & \textcolor[rgb]{1,0,0}{87.4} & \textcolor[rgb]{1,0,0}{99.1} &   \textcolor[rgb]{1,0,0}{99.2} & \textcolor[rgb]{1,0,0}{97.3} & \textcolor[rgb]{1,0,0}{99.0} & \textcolor[rgb]{1,0,0}{84.8} & \textcolor[rgb]{1,0,0}{98.3} & \textcolor[rgb]{1,0,0}{85.8} & \textcolor[rgb]{1,0,0}{99.2} & \textcolor[rgb]{1,0,0}{94.1} & \textcolor[rgb]{1,0,0}{94.8} \\
\hline
\hline
\centering VGG16\&19+SVM~\cite{simonyan2014very} & 99.1 & 89.1 & 96.0 & 94.1 & 74.1 & 92.2 & 85.3 & 97.9 & 79.9 & 92.0 & 83.7 & 97.5 & 96.5 & 94.7 & 97.1 & 63.7 & 93.6 & 75.2 & 97.4 & 87.8 & 89.3 \\
\centering FeV+LV (fusion)~\cite{yang2016exploit} & 98.9 & 93.1 & 96.0 & 94.1 & 76.4 & 93.5 & 90.8 & 97.9 & 80.2 & 92.1 &82.4 & 97.2 & 96.8 & 95.7 & 98.1 & 73.9 & 93.6 & 76.8 & 97.5 & 89.0 & 90.7 \\
\centering HCP+AGS \cite{wei2016hcp,dong2013subcategory} & \textcolor[rgb]{0,0,1}{99.8} & \textcolor[rgb]{0,0,1}{94.8} & 97.7 & 95.4 & 81.3 & 96.0 & 94.5 & 98.9 & 88.5 & 94.1 & 86.0 & 98.1 & 98.3 & 97.3 & 97.3 & 76.1 & 93.9 & 84.2 & 98.2 & 92.7 & 93.2 \\
\centering RCP+AGS \cite{wang2016beyond,dong2013subcategory} & \textcolor[rgb]{0,0,1}{99.8} & 94.5 & \textcolor[rgb]{0,0,1}{98.1} & \textcolor[rgb]{0,0,1}{96.1} & \textcolor[rgb]{0,0,1}{85.5} & \textcolor[rgb]{0,0,1}{96.1} & \textcolor[rgb]{0,0,1}{95.5} & \textcolor[rgb]{0,0,1}{99.0} & \textcolor[rgb]{1,0,0}{90.2} & \textcolor[rgb]{0,0,1}{95.0} & \textcolor[rgb]{0,0,1}{87.8} & \textcolor[rgb]{0,0,1}{98.7} & \textcolor[rgb]{0,0,1}{98.4} & \textcolor[rgb]{0,0,1}{97.5} & \textcolor[rgb]{0,0,1}{99.0} & \textcolor[rgb]{0,0,1}{80.1} & \textcolor[rgb]{0,0,1}{95.9} & \textcolor[rgb]{0,0,1}{86.5} & \textcolor[rgb]{0,0,1}{98.8} & \textcolor[rgb]{0,0,1}{94.6} &  \textcolor[rgb]{0,0,1}{94.3} \\
\centering \textbf{Ours (pre \& fusion)} & \textcolor[rgb]{1,0,0}{99.9} & \textcolor[rgb]{1,0,0}{96.6} & \textcolor[rgb]{1,0,0}{98.4} & \textcolor[rgb]{1,0,0}{97.0} & \textcolor[rgb]{1,0,0}{88.6} & \textcolor[rgb]{1,0,0}{96.4} & \textcolor[rgb]{1,0,0}{95.9} & \textcolor[rgb]{1,0,0}{99.2} & \textcolor[rgb]{0,0,1}{89.0} & \textcolor[rgb]{1,0,0}{97.9} & \textcolor[rgb]{1,0,0}{88.6} &  \textcolor[rgb]{1,0,0}{99.4} & \textcolor[rgb]{1,0,0}{99.3} & \textcolor[rgb]{1,0,0}{97.9} & \textcolor[rgb]{1,0,0}{99.2} & \textcolor[rgb]{1,0,0}{85.8} & \textcolor[rgb]{1,0,0}{98.6} & \textcolor[rgb]{1,0,0}{86.7} & \textcolor[rgb]{1,0,0}{99.4} & \textcolor[rgb]{1,0,0}{95.1} & \textcolor[rgb]{1,0,0}{95.4}\\
\hline
\end{tabular}
\vspace{1pt}
\caption{Comparison of AP and mAP in \% of our model and state-of-the-art methods on the PASCAL VOC 2012 dataset. Upper part presents the results of single model and lower part presents those that aggregate multiple models.
``Ours'' and ``Ours (pre)'' denote our framework without and with pre-training on the COCO dataset. ``Ours (pre \& fusion)'' denotes fusing our two scale results.
The best and second best results are highlighted in {\color{red}{red}} and {\color{blue}{blue}}, respectively. Best viewed in color.}
\label{table:comparision_voc12}
\end{table*}

\subsection{Comparison with State-of-the-art}
To prove the effectiveness of the proposed framework, we conduct extensive experiments on various widely used benchmarks, i.e., Microsoft COCO \cite{lin2014microsoft}, Pascal VOC 2007 \& 2012 \cite{everingham2010pascal}, and Visual Genome \cite{krishna2017visual}.

\subsubsection{Comparison on Microsoft COCO}
Microsoft COCO \cite{lin2014microsoft} is originally constructed for object detection and segmentation, and it has been adopted to evaluate multi-label image classification recently. The dataset contains 122,218 images and covers 80 common categories, which is further divided into a training set of 82,081 images and a validation set of 40,137 images. Since the ground truth annotations of test set are unavailable, our method and all existing competitors are trained on the training set and evaluated on the validation set. For the OP, OR, OF1 and CP, CR, CF1 metrics with top-3 constraint, we follow existing methods \cite{wang2016cnn} to exclude the labels with probabilities lower than a threshold (0.5 in our experiments).

The comparison results are presented in Table \ref{table:result-coco}. As shown, existing best-performing methods are RDAR and ResNet-SRN, in which RDAR adopts a spatial transformer to locate semantic-aware regions and an LSTM network to implicitly capture label dependencies, while ResNet-SRN builds on ResNet-101 and applies attention mechanism to model label relation. The mAP, CF1, and OF1 are 73.4\%, 67.4\%, 72.0\% by RDAR and 77.1\%, 67.4\%, 72.9\% by ResNet-SRN. Different from these methods, our framework incorporates category semantics to better learn semantic-specific feature representations and explores their interactions under the explicit guidance of statistical label co-occurrence, leading to a notable performance improvement on all metrics. Specifically, it achieves the mAP, CF1, and OF1 of 83.8\%, 72.7\%, and 76.2\%, improving those of the previous best methods by 6.7\%, 5.3\%, and 3.3\%, respectively.

\subsubsection{Comparison on Pascal VOC 2007 and 2012}
Pascal VOC 2007 \& 2012 \cite{everingham2010pascal} are the most widely used datasets to evaluate the multi-label image classification task, and most of the existing works report their results on these datasets. Therefore, we conduct experiments on these datasets for more comprehensive comparison. Both datasets cover 20 common categories. Thereinto, Pascal VOC 2007 contains a trainval set of 5,011 images and a test set of 4,952 images, while VOC 2012 consists of 11,540 images as trainval set and 10,991 as test set. For fair comparisons, the proposed framework and existing competitors are all trained on the trainval set and evaluated on the test set.

\begin{figure*}[!t]
   \centering
   \includegraphics[width=0.98\linewidth]{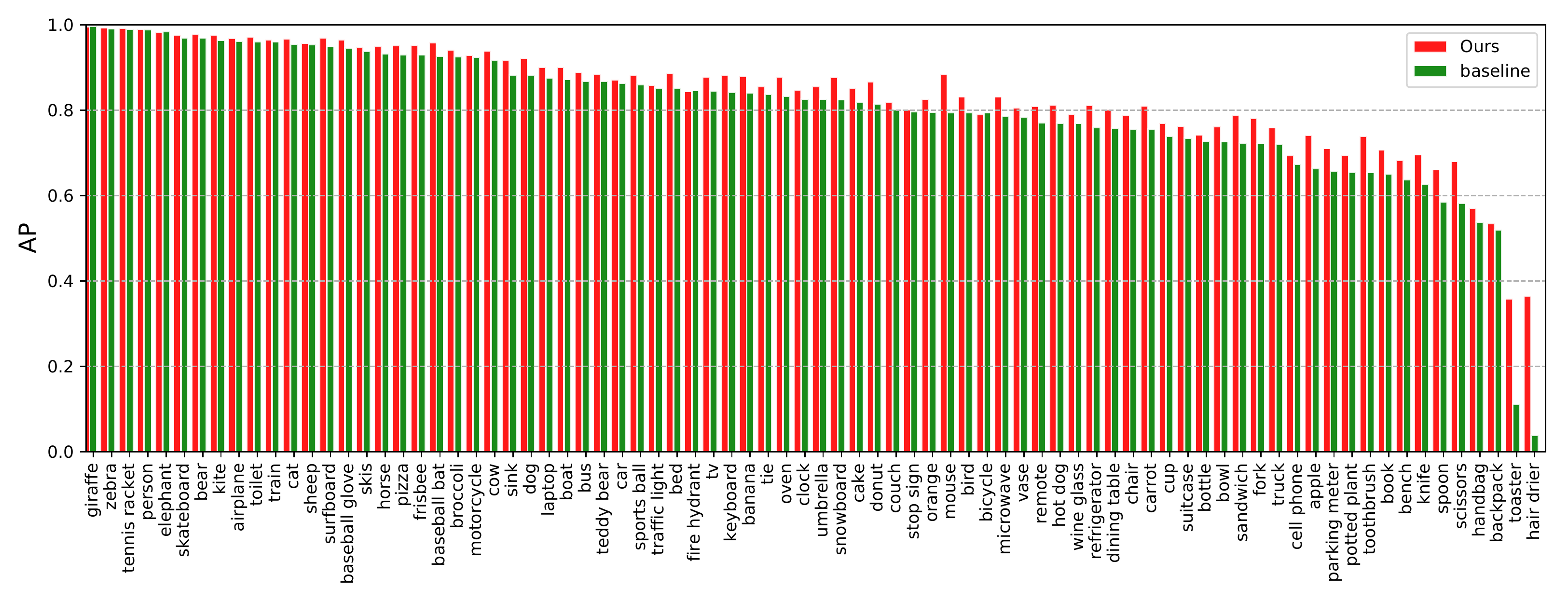} % requires the graphicx package
   \caption{The AP (in \%) of each category of our proposed framework and the ResNet-101 baseline.}
   \label{fig:ap-comparison}
\end{figure*}

We first present the AP of each category and mAP over all categories on the Pascal VOC 2007 dataset in Table \ref{table:comparision_voc07}. Most of existing state-of-the-art methods focus on locating informative regions (e.g., proposal candidates \cite{yang2016exploit,wei2016hcp,zhang2018multi}, attentive regions \cite{wang2017multi}, random regions \cite{wang2016beyond}) to aggregate local discriminative features to facilitate recognizing multiple labels of the given image. For example, RCP achieves a mAP of 92.5\%, which is the best result to date. Differently, our framework incorporates category semantics to better learn semantic-specific features and explores their interactions under the explicit guidance of statistical label dependencies, further improving the mAP to 93.4\%. In addition, by pre-training the framework on the COCO dataset, our framework can obtain an even better performance, i.e., 95.0\% as shown in Table \ref{table:comparision_voc07}. Note that existing methods aggregate multiple models \cite{simonyan2014very} or fuse the result with other methods \cite{yang2016exploit} to improve the overall performance. For example, FeV+LV (fusion) aggregates its results with those of VGG16\&19+SVM, improving the mAP from 90.6\% to 92.0\%. Although our results are generated by a single model, it still outperforms all these aggregated results.

We also compare performance on the Pascal VOC 2012 dataset, as depicted in Table \ref{table:comparision_voc12}. Although VOC 2012 is more challenging and larger in size, our framework still achieves the best performance compared with state-of-the-art competitors. Specifically, it obtains the mAP of 93.9\% and 94.8\% without and with pre-training on the COCO dataset, improving over the previous best method by 1.7\% and 2.6\%, respectively. Similarly, existing methods also aggregate results of multiple models to boost the performance. To ensure fair comparison, we train another model with an input of $448 \times 448$. Specifically, during training, we resize the input image to $512 \times 512$, and we randomly choose a number from {512, 448, 384, 320, 256} as the width and height to randomly crop patches, and further resize the cropped patches to $448 \times 448$. We denote the previous model as scale-640 and this model as scale-512. The two models are all pre-trained on the COCO dataset and retrained on the VOC 2012 dataset. Then, We perform ten crop evaluation (the four corner crops and the center crop as well as their horizontally flipped versions) for each scale and aggregate results from the two scales. As shown in the lower part of Table \ref{table:comparision_voc12}, our framework boosts the mAP to 95.4\%, suppressing all existing methods with single and multiple models.

\subsubsection{Comparison on Visual Genome 500}
Visual Genome \cite{krishna2017visual} is a dataset that contains 108,249 images and covers 80,138 categories. Since most categories have very few samples, we merely consider the 500 most frequent categories, resulting in a VG-500 subset. We randomly select 10,000 images as the test set and the rest 98,249 images as the training set. Compared with existing benchmarks, it covers much more categories, i.e., 500 v.s. 20 on Pascal VOC \cite{everingham2010pascal} and 80 categories on Microsoft-COCO \cite{lin2014microsoft}. To demonstrate the effectiveness of our proposed framework on this dataset, we implement a ResNet-101 baseline network and train it using the same process as ours. As ResNet-SNR \cite{zhu2017learning} is the best-performing method on Microsoft-COCO dataset, we further follow its released code to train ResNet-SNR on this dataset for comparison. All the methods are trained on the training set and evaluated on the test set.

The comparison results are presented in Table \ref{table:result-vg}. Our framework also performs much better than existing state-of-the-art and ResNet-101 baseline methods. Specifically, it achieves the mAP of 36.6\%, improving that of the existing best method by 3.1\%. This comparison clearly suggests that our framework can also work better on recognizing large-scale categories.

\begin{table}[!t]
\centering
\begin{tabular}{c|c}
\hline
\centering  Methods & mAP  \\
\hline
\hline
ResNet-101 \cite{he2016deep} & 30.9\\
ResNet-SRN \cite{zhu2017learning} & 33.5\\
\hline
Ours & 36.6  \\
\hline
\end{tabular}
\vspace{2pt}
\caption{Comparison of mAP (in \%) on the VG-500 dataset.}
\label{table:result-vg}
\end{table}

\subsection{Ablative study}
The proposed framework builds on the ResNet-101 \cite{he2016deep}, thus we compare with this baseline to analyze the contributions of semantic-specific graph representation learning (SSGRL). Specifically, we simply replace the last fully connected layer of the ResNet-101 with a 2,048-to-$C$ fully connected layer and use $C$ sigmoid functions to predict the probability of each category. The training and test settings are exactly the same as those described in Section \ref{sec:optimization}. We conduct experiments on the Microsoft-COCO dataset and present the results in Table \ref{table:result-ablation}. As can be observed, the mAP drops from 83.8\% to 80.3\%. To deeply analyze their performance comparisons, we further present the AP of each category in Figure \ref{fig:ap-comparison}. It shows that the AP improvement is more evident for the categories that are more difficult to recognize (i.e., the categories that the baseline obtains lower AP). For example, for the categories like giraffe and zebra, the baseline obtains very high AP, and our framework just achieves slight improvement. In contrast, for more difficult categories such as toaster and hair drier, our framework improves the AP by a sizeable margin, 24.7\% and 32.5\% improvement for toaster and hair drier, respectively.

\begin{table}[!t]
\centering
\begin{tabular}{c|c}
\hline
\centering  Methods & mAP \\
\hline
\hline
ResNet-101 \cite{he2016deep} &80.3\\
\hline
Ours w/o SD & 80.9  \\
Ours w/o SD-concat & 79.6  \\
Ours w/o SI & 82.2  \\
Ours & 83.8   \\
\hline
\end{tabular}
\vspace{2pt}
\caption{Comparison of mAP (in \%) of our framework (Ours), our framework without SD module (Ours w/o SD and Ours w/o SD-concat) and our framework without SI module (Ours w/o SI) on the Microsoft-COCO dataset.}
\label{table:result-ablation}
\end{table}

The foregoing comparisons verify the contribution of the proposed SSGRL as a whole. Actually, the SSGRL contains two critical modules that work cooperatively, i.e., semantic decoupling (SD) and semantic interaction (SI). In the following, we further conduct ablative experiments to analyze the actual contribution of each module.

\subsubsection{Contribution of semantic decoupling}
We evaluate the contribution of SD module by comparing the performance with and without this module. To this end, we perform average pooling on $\mathbf{f}^I$ to get image feature vector $\mathbf{f}$, and use the following two settings to initialize the graph nodes: 1) directly use $\mathbf{f}$ (namely Ours w/o SD); 2) concatenate $\mathbf{f}$ and corresponding semantic vector (i.e., $\mathbf{x}_c$ for the node corresponding to category $c$), which is mapped to a 2,048 feature vector for initialization (namely Ours w/o SD-concat). As shown in Table \ref{table:result-ablation}, ``Ours w/o SD'' performs slightly better than the baseline method, since it does not incur any additional information but increases the model complexity. ``Ours w/o SD-concat'' performs slightly worse than the baseline and ``Ours w/o SD''. This suggests directly concatenating the semantic vector provide no additional or even interferential information.

\begin{figure}[!t]
   \centering
   \includegraphics[width=1.0\linewidth]{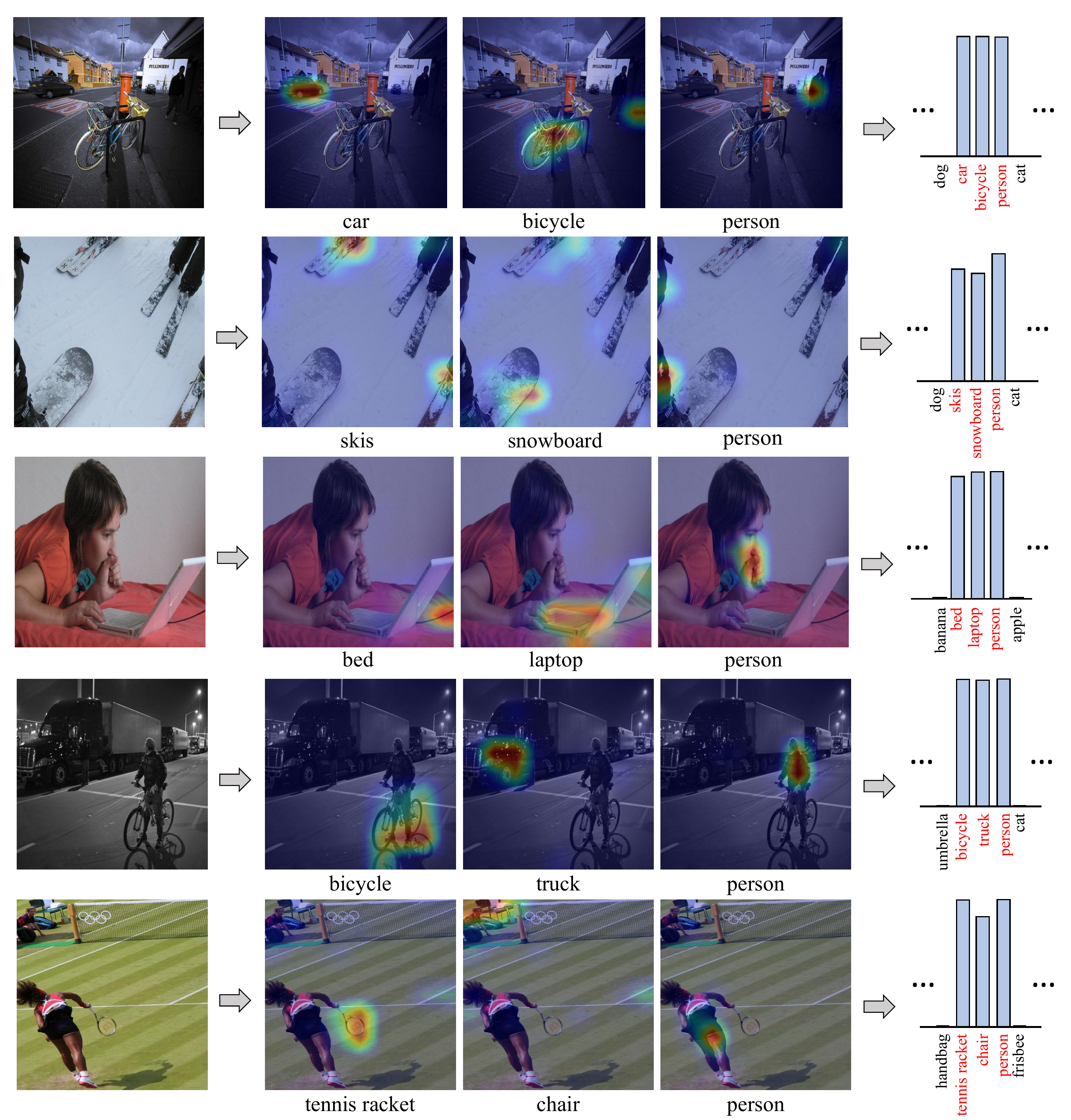} % requires the graphicx package
   \caption{Several examples of input images (left), semantic feature maps corresponding to categories with top 3 highest confidences (middle), and predicted label distribution (right). The ground truth labels are highlighted in red.}
   \label{fig:semantic-map}
   \vspace{-8pt}
\end{figure}

As discussed above, our framework can learn semantic-specific feature maps that focus on corresponding semantic regions via the semantic decoupling. Here, we further visualize some examples in Figure \ref{fig:semantic-map}. In each row, we present the input image, the semantic maps corresponding to categories with the top 3 highest confidences, and the predicted label distribution. It shows that our semantic decoupling module can well highlight the semantic regions if the objects of the corresponding categories exist. For example, the second example has objects of skis, snowboard, and person, our semantic decoupling module highlights the corresponding regions of two skis, snowboard and person leg. Similar phenomena are observed for other examples.

\subsubsection{Contribution of semantic interaction}
To validate the contribution of SI module, we remove the graph propagation network, and thus the classifier $f_c(\cdot)$ directly takes the corresponding decoupled feature vector $\mathbf{f}_c$ as input to predict the probability of category $c$ (namely Ours w/o SI). As shown in Table \ref{table:result-ablation}, we find that its mAP is 82.2\%, decreasing the mAP by 1.6\%.

\section{Conclusion}
In this work, we propose a novel Semantic-Specific Graph Representation Learning framework, in which a semantic guided attentional mechanism is designed to learn semantic-related feature vectors and a graph propagation network is introduced to simultaneously explore interactions among these feature vectors under the guidance of statistical label co-occurrence. Extensive experiments on various benchmarks including Microsoft-COCO, Pascal VOC 2007 \& 2012, and Visual Genome demonstrate the effectiveness of the proposed framework over all existing leading methods.

{\small
\bibliographystyle{ieee_fullname}
\bibliography{egbib}
}

\end{document}